%% file: templateArxiv.tex
\documentclass{article}

\usepackage{PRIMEarxiv}

\usepackage[utf8]{inputenc} 
\usepackage[T1]{fontenc}    
\usepackage{hyperref}       
\usepackage{url}            
\usepackage{booktabs}       
\usepackage{amsfonts}       
\usepackage{nicefrac}       
\usepackage{microtype}      
\usepackage{lipsum}
\usepackage{fancyhdr}       
\usepackage{graphicx}       
\graphicspath{{media/}}     
\usepackage{multirow}
\usepackage{enumitem}
\usepackage[acronym]{glossaries}
\usepackage[dvipsnames]{xcolor}
\usepackage[numbers]{natbib}
\usepackage{adjustbox}
\usepackage[final]{changes}
\usepackage{multirow}
\usepackage[normalem]{ulem}
\useunder{\uline}{\ul}{}

\usepackage[acronym]{glossaries}
\newacronym{gpm}{GPM}{Grid Point Modeling}
\newacronym{vd}{VD}{Voronoi Diagram}
\newacronym{civd}{CIVD}{Clustering Induced Voronoi Diagram}
\newacronym{gnn}{GNN}{Graph Neural Network}
\newacronym{hgt}{HGT}{Heterogeneous Graph Transformer}
\newacronym{mlp}{MLP}{Multi-Layer Perceptron}
\newacronym{gva}{GVA}{Gross Value Added}
\newacronym{osm}{OSM}{OpenStreetMap}
\newacronym{gis}{GIS}{Geographic Information System}
\newacronym{rmse}{RMSE}{Root Mean Squared Error}
\newacronym{kld}{KL-divergence}{Kullback–Leibler divergence}

\usepackage[numbers]{natbib}
\usepackage[normalem]{ulem}
\useunder{\uline}{\ul}{}
\usepackage{tikz}
\usetikzlibrary{shapes.geometric, arrows.meta, positioning, calc, shadows}

\title{Improving Spatial Allocation for Energy System Coupling with Graph Neural Networks
}

\author{
  Xuanhao Mu, Jakob Geiges, Nan Liu, Thorsten Schlachter, Veit Hagenmeyer \\
  Institute for Automation and Applied Informatics (IAI) \\
  Karlsruhe Institute of Technology \\
  76344 Eggenstein-Leopoldshafen, Germany\\
  \{xuanhao.mu, jakob.geiges, nan.liu, thorsten.schlachter, veit.hagenmeyer\}@kit.edu
}

\begin{document}
\maketitle

\begin{abstract}
In energy system analysis, coupling models with mismatched spatial resolutions is a significant challenge.
A common solution is assigning weights to high-resolution geographic units for aggregation, but traditional models are limited by using only a single geospatial attribute. 
This paper presents an innovative method employing a self-supervised Heterogeneous Graph Neural Network to address this issue.
This method models high-resolution geographic units as graph nodes, integrating various geographical features to generate physically meaningful weights for each grid point. 
These weights enhance the conventional Voronoi-based allocation method, allowing it to go beyond simply geographic proximity by incorporating essential geographic information. 
In addition, the self-supervised learning paradigm overcomes the lack of accurate ground-truth data. 
Experimental results demonstrate that applying weights generated by this method to cluster-based Voronoi Diagrams significantly enhances scalability, accuracy, and physical plausibility, while increasing precision compared to traditional methods.
Code is available at \url{https://github.com/KIT-IAI/AllocateGNN}
\end{abstract}


\input{1_introduction}
\input{5_related}
\input{2_method}
\input{3_results}
\input{4_discussion}
\input{6_conclusion}

\bibliographystyle{unsrtnat}
\bibliography{references}

\section*{Acknowledgements}

This work is supported in part by the Helmholtz Association through the project “Helmholtz Platform for the Design of Robust Energy Systems and Their Supply Chains” (RESUR), the Helmholtz Energy System Design (ESD) program, and the Helmholtz AI HAICORE partition (HAICORE@KIT).

\end{document}

%% file: 1_introduction.tex
\section{Introduction}
In the context of global climate change and increasing energy demand, the energy transition and decarbonization have become critical issues. 
Accurate and comprehensive energy system simulation models are essential tools for addressing these challenges. 
The models frequently exhibit varying spatial resolutions, so a key issue in model coupling is the efficient conversion of one model's outputs at a specific spatial granularity into inputs suitable for other models operating at different spatial granularities.
Unlike the traditional spatial disaggregation problem, its ultimate goal is not to generate a high resolution raster map, but to allocate the total amount to a set of nodes or areas with a specific distribution.

\citeauthor{Mu2025Improving}~\cite{Mu2025Improving} used~\gls{gpm} with~\gls{civd} to address the spatial load allocation process, which initially disaggregates the total volume into high resolution intermediate units and subsequently allocates these units to the final target entities utilizing techniques such as the \gls{vd}.
Although this method is superior to traditional~\gls{vd}, it still has the following core limitations in its allocation mechanism:
\begin{enumerate}
    \item Grid point weights are calculated only based on their primary land use type. The grid points of the same type (e.g., all residential points) are given the same weight. 
    \item \gls{gpm} is a static process that struggles to effectively integrate multiple land use proportions or other multi source heterogeneous information. 
\end{enumerate}

The main contributions of this paper are: 
First, we propose a \gls{gpm} framework based on heterogeneous \gls{gnn}, which integrates multiple geospatial features to generate data-driven allocation weights. 
This framework exhibits good scalability and paves the way for the future introduction of more auxiliary data.
Second, we design a self-supervised learning paradigm that uses macroeconomic and socioeconomic indicators as supervisory signals, addressing the key challenge of lacking ground-truth data in the real world. 
Finally, experiments demonstrate that combining this framework with the \gls{civd} method significantly improves spatial allocation accuracy, reducing the \gls{rmse} by an average of $4.87~\%$.
\added{Code is available at \url{https://github.com/KIT-IAI/AllocateGNN}}

This paper is structured as follows: 
Section~\ref{sec:related} presents a review of related work.
Section~\ref{sec:method} outlines the methods used in detail.
Section~\ref{sec:result} describes the data set used in this paper and evaluates the results of the presented methods. 
Section~\ref{sec:discussion} concludes a discussion, and
Section~\ref{sec:conclusion} provides a conclusion and the future work.

%% file: 5_related.tex
\section{Related work}
\label{sec:related}
Energy system simulation models play a crucial role in addressing the dual challenges of global climate change and growing energy demand. 
However, with the increasing prevalence of Distributed Energy Resources (DERs) and the increasing complexity of energy networks, these models generally face granularity gaps~\cite{Cao2021Bridging}. 
Spatial resolution is a core dimension of this gap, directly impacting the reliability and practicality of model results~\cite{Prina2020Classification}.
Improving spatial resolution is crucial for optimizing power system costs and planning~\cite{Frysztacki2021The}, but it must be balanced against computational feasibility \cite{Martinez2021Areview}.


The allocation of macro-regional totals to micro-geographic units is crucial for bridging data across different models.
A classic method is spatial disaggregation, which means breaking spatially aggregated data into smaller spatial units to reveal local features. 
Common spatial disaggregation methods include Areal Weighting~\cite{Qiu2022Disaggregating}, Dasymetric Mapping~\cite{wright1936method}, Machine Learning~\cite{Patil2024A}, and geostatistical methods, such as Kriging Interpolation~\cite{van2004kriging}.

Nonetheless, these traditional spatial disaggregation techniques frequently prove insufficient for directly addressing the granularity gap in model coupling. Their main goal is usually to produce a continuous high-resolution raster map, instead of allocating a total quantity to a defined set of discrete nodes or regions needed by an alternative model. 
\citeauthor{Hülk2017Allocation}~\cite{Hülk2017Allocation} use ~\gls{vd} to address this specific allocation challenge. 
They allocate the annual German electricity consumption and production to the corresponding substations using a two-stage process.
First, using land use data, the total regional volume is spatially disaggregated into several high-spatial-resolution load areas. 
Secondly, the load data is allocated to the corresponding substations.

A~\gls{vd} divides the space into regions, where each region contains all the locations closest to a given point.
Clustering Induced Voronoi Diagram (CIVD)~\cite{Chen2013On} is a novel Voronoi Diagram method that improves the traditional space allocation by dividing spaces using clusters. 
The value of k nearest points for each region in the~\gls{vd} is calculated by clustering.
However, these geometric-based methods inherently rely on pre-defined heuristics (such as distance and proximity) and are unable to dynamically and adaptively learn the complex drivers of spatial allocation.

To incorporate more socioeconomic factors, the method proposed by \citeauthor{Hülk2017Allocation}~\cite{Hülk2017Allocation} utilizes land use data to create a weighted distribution.
\citeauthor{Mu2025Improving}~\cite{Mu2025Improving} used~\gls{gpm} with~\gls{civd} to improve this spatial load allocation process. 
The \gls{gpm} method is first used to calculate the weight distribution based on land use for each grid point, which is obtained by dividing the entire area to be allocated into small squares and has a higher spatial resolution compared to the load area mentioned in \cite{Hülk2017Allocation}. 
Then,~\gls{civd} assigns these grid points to substations through clustering and distance-based rules.
Lastly, the final allocation result (e.g., load distribution) is calculated based on the total weight allocated.
While this method considers more factors than solely geometric methods, its assigned weights are typically calculated statically based on predefined rules.
This complicates the capacity of the model to capture the complex nonlinear relationships between different geographic features.
To overcome these limitations, Graph Neural Networks (GNNs) have been introduced as a powerful framework for learning complex spatial relationships.

\glspl{gnn} are specifically designed to process graph-structured data and can automatically learn complex patterns and relationships from node features and graph topology. 
In recent years, \glspl{gnn} have demonstrated significant potential in a variety of geospatial problems, including urban planning~\cite{Jin2023Graph}, traffic flow prediction~\cite{Bui2021Spatial-temporal}, etc.

By modeling geographic regions and power grid nodes as a heterogeneous graph, \glspl{gnn} are able to learn an end-to-end mapping function. 
Unlike traditional models like Graph Convolutional Networks (GCNs)~\cite{Zhang2019Graph}, which can only process homogeneous nodes, heterogeneous \glspl{gnn}, such as the Heterogeneous Graph Transformer (HGT)~\cite{Hu2020Heterogeneous}, can effectively handle different node types (e.g., administrative districts, substations, and commercial centers) and the rich relationships between them. 
This enables the model to dynamically generate optimal allocation weights for each micro-unit based on multi-source heterogeneous geographical and economic characteristics.

%% file: 2_method.tex
\section{Method}
\label{sec:method}
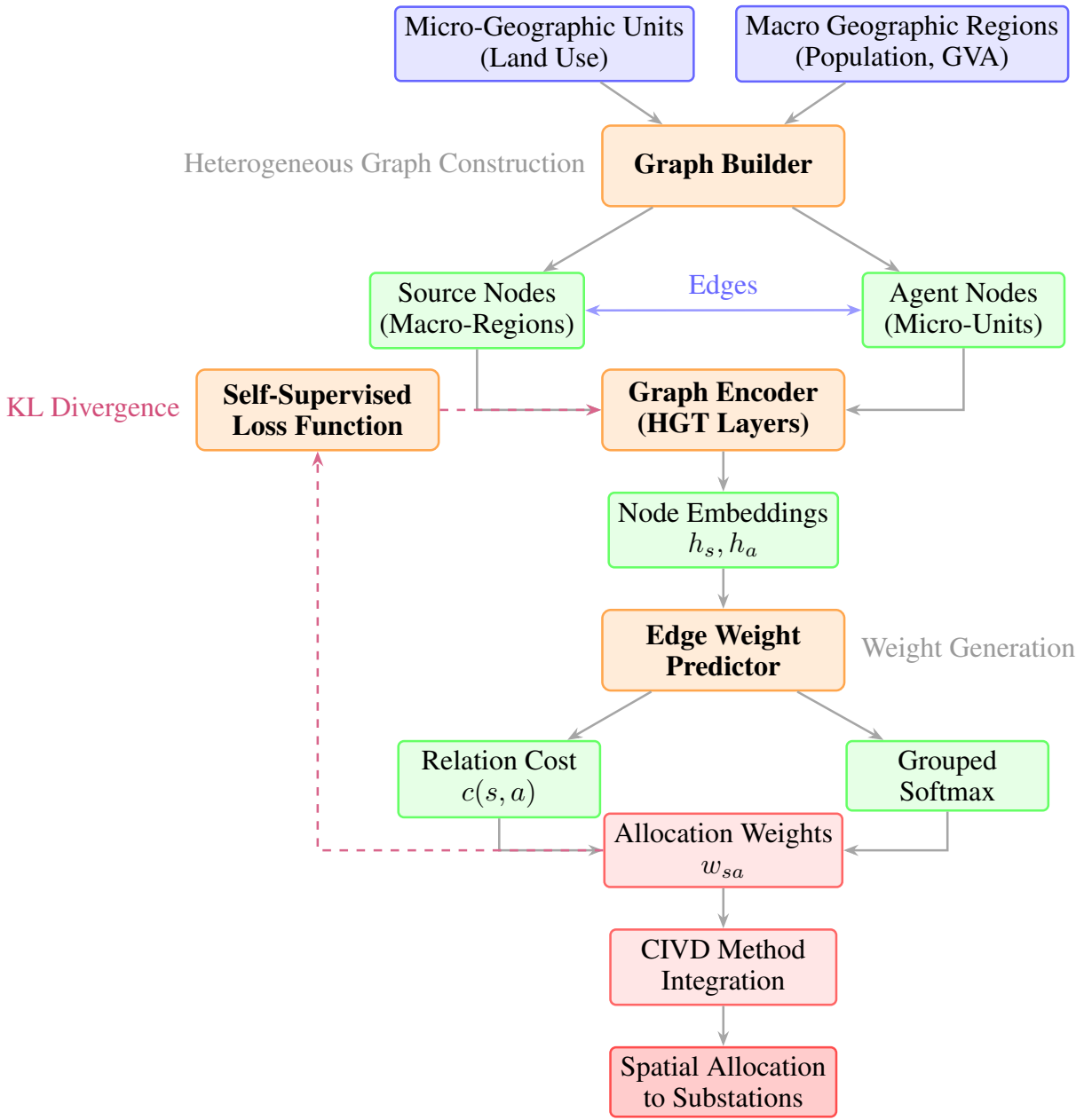
\begin{figure}[htbp]
    \centering
    \resizebox{\columnwidth}{!}{
\begin{tikzpicture}[
    databox/.style={rectangle, draw=blue!60, fill=blue!10, thick, 
                    minimum width=2.8cm, minimum height=0.8cm, 
                    align=center, rounded corners=2pt, font=\normalsize},
    processbox/.style={rectangle, draw=orange!70, fill=orange!15, thick,
                       minimum width=3cm, minimum height=1cm, 
                       align=center, rounded corners=3pt, font=\normalsize\bfseries},
    component/.style={rectangle, draw=green!60, fill=green!10, thick,
                      minimum width=2.5cm, minimum height=0.8cm,
                      align=center, rounded corners=2pt, font=\normalsize},
    output/.style={rectangle, draw=red!60, fill=red!10, thick,
                   minimum width=2.8cm, minimum height=0.8cm,
                   align=center, rounded corners=2pt, font=\normalsize},
    arrow/.style={-{Stealth[scale=0.8]}, thick, color=gray!70},
    doublearrow/.style={{Stealth[scale=0.8]}-{Stealth[scale=0.8]}, thick, color=gray!70},
    label/.style={font=\normalsize, color=gray!80}
]

\node[databox] (geo) {Micro-Geographic Units\\(Land Use)};
\node[databox, right=0.5cm of geo] (macro) {Macro Geographic Regions\\(Population, GVA)};

\node[processbox, below=1cm of $(geo)!0.5!(macro)$] (builder) {Graph Builder};

\node[component, below left=0.8cm and 0.2cm of builder] (source) {Source Nodes\\(Macro-Regions)};
\node[component, below right=0.8cm and 0.2cm of builder] (agent) {Agent Nodes\\(Micro-Units)};

\node[processbox, below=2cm of builder] (encoder) {Graph Encoder\\(HGT Layers)};

\node[component, below=0.5cm of encoder] (embeddings) {Node Embeddings\\$h_s, h_a$};

\node[processbox, below=0.5cm of embeddings] (predictor) {Edge Weight\\Predictor};

\node[component, below left=0.6cm and 0cm of predictor, font=\normalsize] (cost) {Relation Cost\\$c(s,a)$};
\node[component, below right=0.6cm and 0cm of predictor, font=\normalsize] (softmax) {Grouped\\Softmax};

\node[output, below=1.5cm of predictor] (weights) {Allocation Weights\\$w_{sa}$};

\node[processbox, left=2cm of encoder] (loss) {Self-Supervised\\Loss Function};

\node[output, below=0.5cm of weights] (civd) {CIVD Method\\Integration};

\node[output, below=0.5cm of civd, fill=red!20, draw=red!70] (final) {Spatial Allocation\\to Substations};

\draw[arrow] (geo) -- (builder);
\draw[arrow] (macro) -- (builder);
\draw[arrow] (builder) -- (source);
\draw[arrow] (builder) -- (agent);
\draw[arrow] (source) |- (encoder);
\draw[arrow] (agent) |- (encoder);
\draw[arrow] (encoder) -- (embeddings);
\draw[arrow] (embeddings) -- (predictor);
\draw[arrow] (predictor) -- (cost);
\draw[arrow] (predictor) -- (softmax);
\draw[arrow] (cost) |- (weights);
\draw[arrow] (softmax) |- (weights);
\draw[arrow] (weights) -- (civd);
\draw[arrow] (civd) -- (final);

\draw[arrow, dashed, color=purple!60] (weights) -| (loss);
\draw[arrow, dashed, color=purple!60] (loss) -- (encoder);

\node[label, left=2pt of builder, anchor=east] {Heterogeneous Graph Construction};
\node[label, right=2pt of predictor, anchor=west] {Weight Generation};
\node[label, left=2pt of loss, anchor=east, color=purple!70] {KL Divergence};

\draw[doublearrow, color=blue!40] (source) -- (agent) node[midway, above, font=\normalsize, color=blue!60] {Edges};

\end{tikzpicture}
}

\caption{Flowchart of the GNN-based spatial allocation method. The main pipeline (blue) generates weights, while the self-supervised loop (red) computes the loss for training by reconstructing macro-indicators.}
\label{fig:method_flowchart}
\end{figure}

The present paper proposes a Graph Neural Network (GNN)-based Grid Point Method (GPM) to generate allocation weights for grid points in geographic space. 
The core idea is to model the spatial allocation problem as a graph learning task. 
The overall pipeline consists of four core components: Graph Builder, Graph Encoder, Edge Weight Predictor, and Loss Function, as shown in Fig.~\ref{fig:method_flowchart}.

\subsection{Graph Builder}
As the task involves two functionally distinct spatial entities, macro-regions (source nodes) and micro-geographic units (agent nodes), with an asymmetric allocation relationship, we construct a heterogeneous graph to explicitly model this relationship.
The graph consists of two different types of nodes:
\begin{itemize}
    \item Source Nodes $\mathcal{N}(s)$: These represent the macro-geographic regions whose total volume (e.g., the region's annual total electricity consumption) needs to be decomposed and allocated.
    Node features are primarily derived from macro-statistics of the region, such as total population and~\gls{gva} of various industries, to describe the socioeconomic profile of the region.
    \item Agent Nodes $\mathcal{N}(a)$: These represent the micro-geographic units that receive the allocated volume. 
    Uniform, fixed size grids are generated across the entire macro-geographic region, using the center of each grid as an agent node. 
    This discretization method transforms the continuous geographic space into a manageable set of nodes.
    Node features are primarily derived from geographic information data, such as~\gls{osm}. 
    Specifically, including the area percentage of different land use types (e.g., residential and industrial) within the grid, as well as a One-Hot encoding of the dominant land use type.
\end{itemize}
The connections between nodes, named edges, are crucial to defining how information flows within the graph.
In the present paper, edges are constructed from direct geographic affiliations. 
If the geographic coordinates of an agent node are within the polygonal boundary of a source node, a connection is established between them. 
To enable bidirectional information exchange, we establish two directed edges for each pair of connected nodes: one from the source node to the agent node, representing the flow of macro-region information to micro-units; and another from the agent node back to the source node, allowing the model to aggregate the states of micro-units to update its understanding of the macro-region.

\subsection{Graph Encoder}
The core task of the Graph Encoder is to learn an embedding vector for each node in the graph. 
This task faces two major challenges: the original features of the source and agent nodes are heterogeneous, with different dimensions, types, and semantic spaces; the complex relationships defined by different node types and edges in the graph need to be effectively captured. 
To address these challenges, the encoder adopts a two-stage architecture. 
First, separate linear layers project the original features of source and agent nodes into a unified latent space. 
Subsequently, we use a multi-layer~\gls{hgt} as the core information aggregation module.
\gls{hgt} is an attention architecture explicitly designed for heterogeneous graphs and can learn unique semantic weights for each relationship based on the types of nodes and edges.

Finally, the encoder generates an embedding vector for each node that combines its own characteristics with graph structure information, providing high quality input for downstream weight prediction tasks.

\subsection{Edge Weight Predictor}
After the Graph Encoder generates node embeddings, the Edge Weight Predictor converts these abstract representations into concrete assigned weights.
This fully differentiable prediction module aims to predict a weight $w_{sa}$ for each edge connecting a source node $s \in \mathcal{N}(s)$ with an agent node $a \in \mathcal{N}(a)$, ensuring that the sum of the weights of all edges originating from the same source node is strictly 1, i.e. $\sum_{a \in \mathcal{N}(s)} w_{sa} = 1$.
This module includes Relation Cost Calculation and Grouped Softmax Normalization.

In the embedding space, an agent node should be semantically closer to the source node to which it belongs. 
This relationship cost $c(s,a)$ can be quantified by a gated distance. 
It is obtained by multiplying two parts:
\begin{itemize}
    \item Euclidean Distance: Directly computes the distance $d(s,a)$ between the source node embedding $h_s$ and the agent node embedding $h_a$.
    \item Gating Score: A small~\gls{mlp} takes the concatenated vector of the two node embeddings $[h_s || h_a]$ as input and outputs a relationship score $g(s,a)$. This score serves as a learnable gating signal to adjust the importance of distance dynamically.
\end{itemize}
The final relationship cost is calculated as follows: 
\begin{equation}
    c(s, a) = \underbrace{\sigma(\text{MLP}([h_s || h_a]))}_{g(s, a): \text{Gating Score}} \ \ \ \cdot \underbrace{\|h_s - h_a\|_2}_{d(s, a): \text{Euclidean Distance}}
\end{equation}
where $[h_s || h_a]$ represents vector concatenation, and $\sigma$ is a sigmoid function that ensures the gating score is between $(0, 1)$. 
Lower cost values $c(s,a)$ indicate a better match between agent node $a$ and source node $s$.

In order to convert the "lower is better" cost value into a "higher is better" weight that "sums to 1", it is necessary to apply the Softmax function to the negative costs of all outgoing edges of each source node $s$:
\begin{equation}
    w_{sa} = \frac{\exp(-c(s, a) / \tau)}{\sum_{a' \in \mathcal{N}(s)} \exp(-c(s, a') / \tau)}
\end{equation}
where $a' \in \mathcal{N}(s)$ is the set of all agent nodes connected to the source node s. $\tau$ is a temperature hyperparameter that controls the smoothness of the weight distribution.

Through the above design, the entire edge weight predictor is fully differentiable. This ensures that the gradient of the loss can be smoothly back-propagated to each layer of the graph encoder.

\subsection{Loss Function}
One of the primary challenges in spatial allocation is the lack of ground-truth data. 
This means that for consumption allocation problems, although some aggregated load data may exist at the substation level, the precise and true load values for each individual micro-geographic unit are usually unavailable.
To overcome this bottleneck, a self-supervised learning framework is used. 
Instead of relying on direct labels, this framework generates supervisory signals by defining agent tasks based on the intrinsic structure of the data and domain knowledge, thereby indirectly but effectively guiding model training.

The core loss function is the land use prediction loss. 
This loss function stems from a core domain assumption: the~\gls{gva} and population share of different functional areas (e.g., industrial and commercial areas) within a macro-region should be highly correlated with the distribution of energy consumption across these functional categories~\cite{Hülk2017Allocation}.
Therefore, if the distribution weights $w_{sa}$ learned by the model are meaningful, it should be able to reconstruct the macro-distribution of socioeconomic indicators by aggregating the attributes of agent nodes $\mathcal{N}(a)$.

First, each source node $s$ constructs a true, normalized distribution vector $P_s$.
\begin{equation}
    P_s = \left[ \frac{\text{GVA}_{\text{industry}}}{\text{Total}}, \frac{\text{GVA}_{\text{commerce}}}{\text{Total}}, ..., \frac{\text{Population}}{\text{Total}} \right]_s
\end{equation}

Then, the model uses its predicted edge weights $w_{sa}$ to perform a weighted summation of the land use type of each agent node $a$ (represented as a One-Hot vector $T_a$) to reconstruct the distribution $\hat{P}_s$.
\begin{equation}
    \hat{P}_s = \sum_{a \in \mathcal{N}(s)} w_{sa} T_a
\end{equation}

Finally, \gls{kld} is used to measure the difference between the reconstructed distribution and the true distribution P. The total loss is the sum of \gls{kld} over all source nodes:
\begin{equation}
    \mathcal{L}_{\text{self-supervised}} = \sum_{s \in \mathcal{S}} D_{KL}(P_s \ || \ \hat{P}_s)
\end{equation}

\subsection{Clustering Introduced Voronoi Diagram}
The \gls{gnn} model proposed in this paper serves as a core weight generation module and is integrated with the \gls{civd} method~\cite{Mu2025Improving}. 
This constitutes a two-stage spatial allocation process, enabling allocation from macro-regional aggregate demand to specific power facilities.
The weighted agent nodes generated by \gls{gnn} will serve as input to the \gls{civd} allocation process.
The \gls{civd} method first clusters power facilities (e.g., substations) in geographic space to form multiple load centers. 
Then, for each agent node, the algorithm assigns it to the nearest load center based on its Euclidean distance to each load center.
Finally, the weighted agent nodes within each cluster are aggregated and evenly distributed to power facilities within that region.

For the implementation and theoretical basis of the \gls{civd} method, please refer to research~\cite{Mu2025Improving} and~\cite{Chen2013On}.

%% file: 3_results.tex
\section{Results}
\label{sec:result}

\begin{table*}[!ht]
\footnotesize
\renewcommand{\arraystretch}{1.3}
\centering
\caption{Comparison of Root Mean Squared Error (RMSE) Performance Across Various Geographic Regions}
\label{table_rmse}
\begin{tabular}{|lrrrrrr|}
\hline
\multicolumn{1}{|c|}{Region} & \multicolumn{1}{c|}{VD} & \multicolumn{1}{c|}{VD-GPM} & \multicolumn{1}{c|}{VD-GNN-GPM} & \multicolumn{1}{c|}{CIVD} & \multicolumn{1}{c|}{CIVD-GPM} & \multicolumn{1}{c|}{CIVD-GNN-GPM} \\ \hline
\multicolumn{7}{|c|}{Training Set} \\ \hline
\multicolumn{1}{|l|}{London} & 21.085 & 16.234 & \multicolumn{1}{r|}{17.013 (\underline{-4.80\%})} & 19.933 & 16.220 & \textbf{15.998 (+1.37\%)} \\ \hline
\multicolumn{1}{|l|}{TLH2} & 8.805 & 6.735 & \multicolumn{1}{r|}{6.656 (+1.17\%)} & 7.647 & 5.816 & \textbf{5.778 (+0.65\%)} \\ \hline
\multicolumn{1}{|l|}{TLH3} & 9.269 & 7.265 & \multicolumn{1}{r|}{7.064 (+2.77\%)} & 9.069 & 6.727 & \textbf{6.535 (+2.85\%)} \\ \hline
\multicolumn{1}{|l|}{TLJ1} & 15.831 & 9.981 & \multicolumn{1}{r|}{9.979 (+0.02\%)} & 13.676 & 8.917 & \textbf{8.493 (+4.76\%)} \\ \hline
\multicolumn{1}{|l|}{TLF1} & 11.966 & 8.591 & \multicolumn{1}{r|}{8.129 (+5.38\%)} & 10.840 & 7.627 & \textbf{7.072 (+7.28\%)} \\ \hline
\multicolumn{1}{|l|}{TLF2} & 12.574 & 9.301 & \multicolumn{1}{r|}{9.266 (+0.38\%)} & 10.884 & 7.542 & \textbf{6.970 (+7.58\%)} \\ \hline
\multicolumn{1}{|l|}{TLC1} & 18.302 & 9.965 & \multicolumn{1}{r|}{9.494 (+4.73\%)} & 17.630 & 8.882 & \textbf{7.873 (+11.36\%)} \\ \hline
\multicolumn{1}{|l|}{TLC2} & 13.318 & 9.936 & \multicolumn{1}{r|}{9.484 (+4.55\%)} & 8.173 & 5.841 & \textbf{5.820 (+0.36\%)} \\ \hline
\multicolumn{1}{|l|}{TLD6} & 8.674 & 6.139 & \multicolumn{1}{r|}{5.991 (+2.41\%)} & 7.769 & 5.384 & \textbf{5.017 (+6.82\%)} \\ \hline
\multicolumn{1}{|l|}{TLG1} & 15.492 & 11.041 & \multicolumn{1}{r|}{10.861 (+1.63\%)} & 14.763 & 10.740 & \textbf{9.983 (+7.05\%)} \\ \hline
\multicolumn{1}{|l|}{TLG2} & 13.909 & 9.563 & \multicolumn{1}{r|}{\textbf{9.032 (+5.55\%)}} & 12.984 & 11.795 & 11.694 (+0.86\%) \\ \hline
\multicolumn{1}{|l|}{TLE4} & 13.770 & 7.449 & \multicolumn{1}{r|}{7.430 (+0.26\%)} & 13.353 & 7.007 & \textbf{6.688 (+4.55\%)} \\ \hline
\multicolumn{7}{|c|}{Test Set} \\ \hline
\multicolumn{1}{|l|}{TLH1} & 7.532 & 5.514 & \multicolumn{1}{r|}{5.636 (\underline{-2.21\%})} & 7.188 & 5.670 & \textbf{5.407 (+4.64\%)} \\ \hline
\multicolumn{1}{|l|}{TLE3} & 17.336 & 9.357 & \multicolumn{1}{r|}{9.521 (\underline{-1.75\%})} & 12.720 & 7.479 & \textbf{6.870 (+8.14\%)} \\ \hline
\multicolumn{1}{|l|}{TLD3} & 10.201 & 7.068 & \multicolumn{1}{r|}{7.153 (\underline{-1.20\%})} & 9.592 & 6.579 & \textbf{6.266 (+4.76\%)} \\ \hline
\multicolumn{1}{|l|}{TLD4} & 11.537 & 5.636 & \multicolumn{1}{r|}{5.830 (\underline{-3.44\%})} & 9.808 & \textbf{4.424} & 4.443 (\underline{-0.43\%}) \\ \hline
\end{tabular}
\end{table*}

The present paper uses the Great Britain primary substations dataset from~\cite{ZHOU2024Dataset} to verify the effectiveness of the described method. 
This dataset contains the peak demand at $11~kV$ substations in various main regions of Great Britain.
The experiments used data from 16 different International Territorial Level (ITL)~\cite{uk_itl} regions. 
The train set included 12 regions: London (TLI3-7), TLH2 (Cheshire), TLH3 (Greater Manchester), TLJ1 (West Midlands), TLF1 (Derbyshire and Nottinghamshire), TLF2 (Leicestershire, Rutland and Northamptonshire), TLC1 (Cumbria), TLC2 (Lancashire), TLD6 (South Yorkshire), TLG1 (Herefordshire, Worcestershire and Warwickshire), TLG2 (Staffordshire and Shropshire) and TLE4 (North Northamptonshire). The test set included the last 4 regions not included in training: TLH1 (Merseyside), TLE3 (Lincolnshire), TLD3 (South Yorkshire), and TLD4 (West Yorkshire).
Due to its ultra-dense nature, London is evaluated as a separate region, which can be used to test the model's performance in extreme cases.
Each region is divided into approximately $50,000$ grid points at different grid resolutions.
All calculations in this paper were executed using Python.
Geographic data was processed and analyzed with the GeoPandas Python library~\cite{jordahl_geopandasgeopandas_2020}. 
The~\gls{gis} data is sourced from~\gls{osm}~\cite{openstreetmap_contributors_planet_2017} and is retrieved using the OSMnx~\cite{boeing_modeling_2024} Python library.
Population statistics and~\gls{gva} figures for administrative regions were obtained from sources~\cite{uk_population} and~\cite{uk_gva}.

\subsection{Allocation Performance}

Tab.~\ref{table_rmse} compares the performance of our proposed method with several baseline methods across various geographic regions, using the \gls{rmse} as the evaluation metric. 
Lower \gls{rmse} values indicate superior model performance. 
The abbreviations for the methods in the table are as follows: VD (Voronoi Diagram), GPM (Grid Point Modeling), and CIVD (Cluster Introduced Voronoi Diagram). 
To clearly present the comparison results, bold numbers in the table indicate the best performance among all methods in that region.
The percentages in parentheses quantify the performance improvement of the \gls{gnn} for \gls{gpm}, representing the percentage reduction achieved by the \gls{gnn} enhancement method (e.g., CIVD-GNN-GPM) relative to its corresponding \gls{gpm} baseline (e.g., CIVD-GPM).
A positive value represents a performance optimization (RMSE reduction), while a negative value indicates performance degradation (RMSE increase, indicated with an underline).

The model demonstrated good learning and fitting capabilities on the training set. 
The CIVD-GNN-GPM method consistently outperforms the previous CIVD-GPM~\cite{Mu2025Improving} method without the GNN across all training regions, achieving an average RMSE reduction of $4.87\%$. 
For example, in the TLC1 region, the application of GNN-GPM reduced the \gls{rmse} from $8.882$ to $7.873$, corresponding to an error reduction of $11.36~\%$. 
This steady improvement is significant because the GNN model doesn't use any real substation data as a supervisory signal. 
Furthermore, the training was based solely on open-source data from \gls{osm}.
It demonstrates that the model can successfully learn physically meaningful and realistic load spatial distribution patterns based solely on the distribution patterns of macroeconomic and socioeconomic indicators.
This improvement is not only a statistical optimization but also represents a significant improvement in the fidelity of the load distribution model. 
This also lays a good foundation for the future introduction of additional auxiliary data.

However, the performance of the overall two-stage allocation framework hinges on the compatibility of the methodologies between modules. 
This is exemplified by comparing GNN-enhanced methods with traditional Voronoi Diagram (VD) baselines.
For example, the CIVD-GNN method almost always outperforms CIVD. 
However, in some areas, VD-GPM underperforms VD.
For example, in London, the application of GNN-GPM improves the \gls{rmse} from  $17.013$ to $16.234$.
This phenomenon can be attributed to a synergistic mismatch between the weight generation and spatial allocation stages of the two-stage allocation method. 
GPM can generate a weight distribution that is close to the actual demand. However, traditional VD strictly follows geometric partitioning, and its rigid partition boundaries may cause spatial misalignment with the high-load areas identified by GNN.
This misalignment can lead to error amplification: a high-load area identified by \gls{gnn} may be completely misattributed to a suboptimal facility by the VD boundary, resulting in a larger error than the fuzzy average allocation.
This comparison proves that the overall performance of the hybrid model depends not only on the advancement of individual modules, but more importantly on the compatibility of the methodologies between modules.

\begin{figure}[!ht]
\centering
\includegraphics[width=\columnwidth]{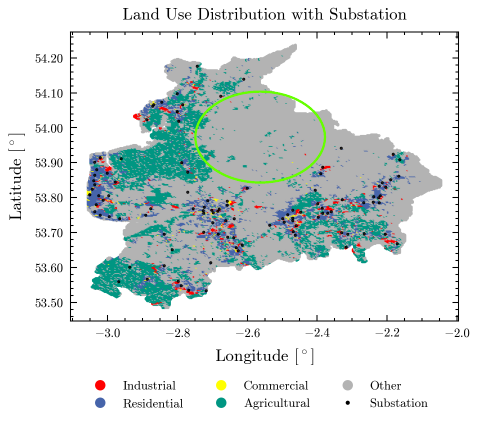}
\caption{Land Use Distribution with Substation in Region TLD4}
\label{fig_tld4}
\end{figure}

This compatibility issue is also reflected in the test set.
The CIVD-GNN-GPM approach maintains its performance advantage when applied to four geographic regions not included in the training set. 
Specifically, this approach achieves lower \gls{rmse} values than CIVD-GPM across three test regions, except TLD4. 
Numerically, TLD4's performance difference is 0.019, representing a $0.43~\%$ increase in \gls{rmse}. 
Fig.~\ref{fig_tld4} illustrates that the upper part of TLD4 (shown by the green circle) comprises a substantial ``other'' region devoid of a substation at its center. 
This region probably utilizes no electricity; nonetheless, both the GNN model and the GPM approach attribute weights to it. 
This is because neither method knows the exact location of the substation.
However, GNNs, by learning the different attributes of grid points, produce a sharper weight distribution than GPM. This sharp weight distribution, combined with the rigid distribution, amplifies the error, leading to performance degradation.
It highlights the need to improve the compatibility of hybrid models in the absence of real, labeled data.

\subsection{Physical Plausibility}

\begin{figure*}[!ht]
\centering
\includegraphics[width=\textwidth]{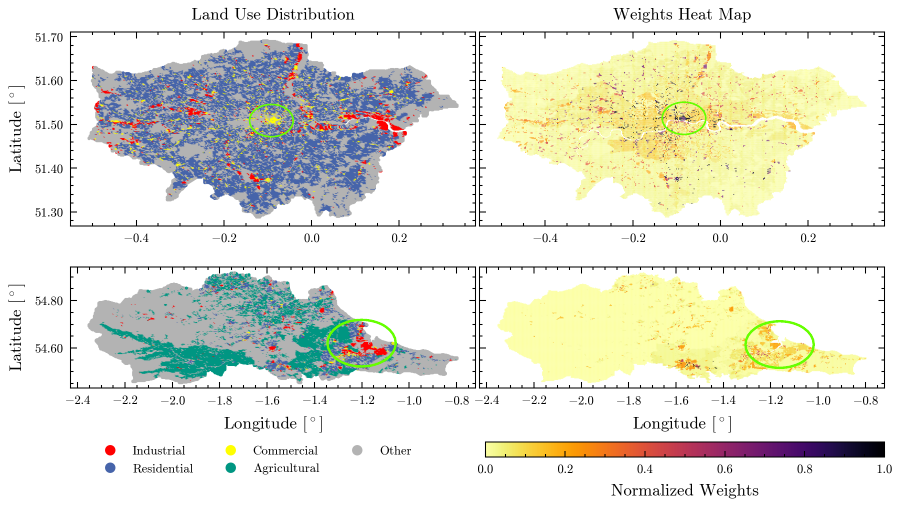}
\caption{Land Use Distribution and Weights Heat Map for region London (top) and TLC1 (bottom)}
\label{fig_sim}
\end{figure*}

To verify the physical plausibility of the proposed method, Fig.~\ref{fig_sim} qualitatively demonstrates the land use distribution and weight distribution generated by the GNN for two different geographic regions: London (top) and the TLC1 region (bottom).
Using London as an example, the heat map (top right) shows a high concentration of weights in the urban core (green circle), which is consistent with the high density of commercial (yellow) and residential (blue) areas shown in the land use map (top left). 
Moreover, for the TLC1 region (bottom), which contains large tracts of agricultural land (cyan), the model assigns significantly higher weights to the eastern region (green circle), which is home to a concentration of industrial (red) and residential (blue) areas. 
In contrast, the weights for the vast agricultural areas and other undeveloped areas are negligible.
Overall, these visualizations demonstrate the model's ability to learn the complex relationship between geospatial characteristics and energy demand, resulting in a spatially consistent and physically plausible allocation scheme.

%% file: 4_discussion.tex
\section{Discussion}
\label{sec:discussion}

The results of this study clearly demonstrate that combining \gls{gnn} with \gls{civd} can significantly improve the accuracy of spatial allocation of energy systems. 
By abstracting geographic space into a heterogeneous graph, \glspl{gnn} are able to capture the complex, nonlinear relationships between multidimensional features such as land use, population density, and economic activity, thereby generating dynamic, data-driven allocation weights for each micro-geographic unit.
Furthermore, compared to the static GPM approach, GNN can incorporate more auxiliary data through node features to dynamically calculate weights, demonstrating good scalability.
Moreover, the model training does not require any ground-truth data and is entirely self-supervised.

While the proposed approach has achieved positive results, it still has some limitations. 
First, the model's performance is highly dependent on the quality and granularity of the input data, such as the coverage and accuracy of \gls{osm} data. 
Its effectiveness may be limited in data-sparse areas. 
Subsequently, there are performance differences when \gls{gnn} is combined with different spatial delineation methods.
In the TLD4 region, the GNN, due to its lack of the locations of substations, produced a more concentrated weight distribution in an anomalous area lacking a substation.
In this case, higher accuracy became a disadvantage, amplifying the overall error.
This reveals the need to prioritize compatibility between modules when designing a multi-stage framework, rather than simply superimposing advanced technologies.

%% file: 6_conclusion.tex
\section{Conclusion}
\label{sec:conclusion}
To address the critical spatial granularity gap in coupled energy systems, this paper proposes an innovative two-stage spatial allocation framework based on Graph Neural Network (GNN). 
The key contribution of this method lies in the first application of heterogeneous \gls{gnn} to Grid Point Modeling (GPM), which is a method for generating high-resolution spatial distribution weights based on geographic features. 
A self-supervised learning paradigm is designed to enable this method to generate physically reasonable allocation weights by learning from the distribution patterns of macro-socioeconomic indicators, even in the absence of ground-truth data.
Furthermore, the node features in the \gls{gnn} provide good scalability for the future introduction of additional auxiliary data.

Experimental results validate the effectiveness of this approach. By combining the dynamic weights generated by the \gls{gnn} with Clustering Induced Voronoi Diagram (CIVD), the proposed CIVD-GNN-GPM model outperforms previous methods in accuracy and demonstrates good generalization on unseen test data. 
This demonstrates that this framework not only improves allocation accuracy but also provides a new and practical method for solving spatial allocation problems. 
\added{However, our self-supervised objective relies on macro-proxies to supervise micro-level demand allocation, which can be biased when electricity use decouples from these indicators (e.g., energy-intensive industrial loads or data centers).
Future work will integrate energy-intensity proxies and uncertainty-aware weighting to improve robustness under proxy mismatch and mitigate systematic allocation errors.}
\deleted{Future research could focus on integrating geographic information of target nodes (e.g., substations) directly into the \gls{gnn} model. 
This aims to enable the model to autonomously identify geographically outliers (e.g., large areas lacking substations), thereby correcting allocation biases and improving accuracy while building an end-to-end allocation framework.}

\section{Declaration}
During the preparation of this work, the authors used Gemini (Google DeepMind) and Claude (Anthropic) in order to perform text polishing and figure creation. After using this tool/service, the author(s) reviewed and edited the content as needed and take full responsibility for the content of the publication.